\documentclass[10pt,twocolumn,letterpaper]{article}

\usepackage{cvpr}              

\usepackage{multirow}
\usepackage[utf8]{inputenc} 
\usepackage[T1]{fontenc}    
\usepackage{graphicx}
\usepackage{newunicodechar}
\newunicodechar{，}{,}
\usepackage{url}            
\usepackage{booktabs}       
\usepackage{amsfonts}       
\usepackage{nicefrac}       
\usepackage{microtype}      
\usepackage{xcolor}         
\usepackage{bbding}
\usepackage{amsmath}
\usepackage[accsupp]{axessibility} 
\usepackage{mathtools}
\usepackage{amsthm}
\usepackage{amssymb}
\usepackage{bm}
\usepackage{tcolorbox}
\usepackage{xcolor} 
\usepackage{color} 
\usepackage{soul}
\usepackage{subcaption}
\usepackage{colortbl}
\usepackage{wrapfig}
\definecolor{color1}{RGB}{135,206,250}

\definecolor{color2}{RGB}{255,228,225}
\def\logo{\makebox[22pt][l]{\raisebox{-0.9ex}{\includegraphics[height=20pt]{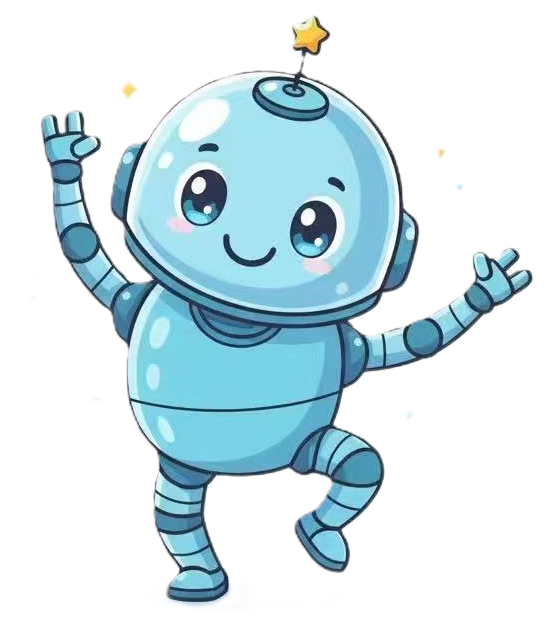}}\hspace{15pt}}}
\definecolor{cvprblue}{rgb}{0.21,0.49,0.74}
\usepackage[pagebackref,breaklinks,colorlinks,allcolors=cvprblue]{hyperref}

\def\paperID{1279} 
\def\confName{CVPR}
\def\confYear{2026}

\title{Do You Have Freestyle? Expressive Humanoid Locomotion via Audio Control}

\author{
Zhe Li$^{1,2, \S}$\thanks{Equal Contribution ~~~ $\dagger$ Corresponding Author} ,
Yangyang Wei$^{2,4 *}$,
Boan Zhu$^{2,5 *}$,
Tao Huang$^{6}$,
Zhenguo Sun$^{2}$,
Yibo Peng$^{2}$,\\
~Pengwei Wang$^{2}$,
Zhongyuan Wang$^{2}$,
Fangzhou Liu$^{4}$,
Chang Xu$^{3}$,
Cheng Chi$^{2 \dagger}$,
Shanghang Zhang$^{7 \dagger}$ \\
$^{1}$ Nanyang Technological University, $^{2}$ BAAI, $^{3}$ University of Sydney, $^{4}$ Harbin Institute of Technology\\
$^{5}$ Hong Kong University of Science and Technology,
$^{6}$ Shanghai Jiao Tong University\\
$^{7}$ State Key Laboratory of Multimedia Information Processing, Peking University \\
}
\begin{document}
\twocolumn[{%
\renewcommand\twocolumn[1][]{#1}%
\maketitle
\vspace{-13mm}
\begin{center}
\small
$^{\star}$Equal Contribution,
$^{\dagger}$Corresponding Author,
$\S$ Project Leader
\end{center}
\begin{center}
    \captionsetup{type=figure}
    \includegraphics[width=0.9\linewidth,height=0.5\linewidth]{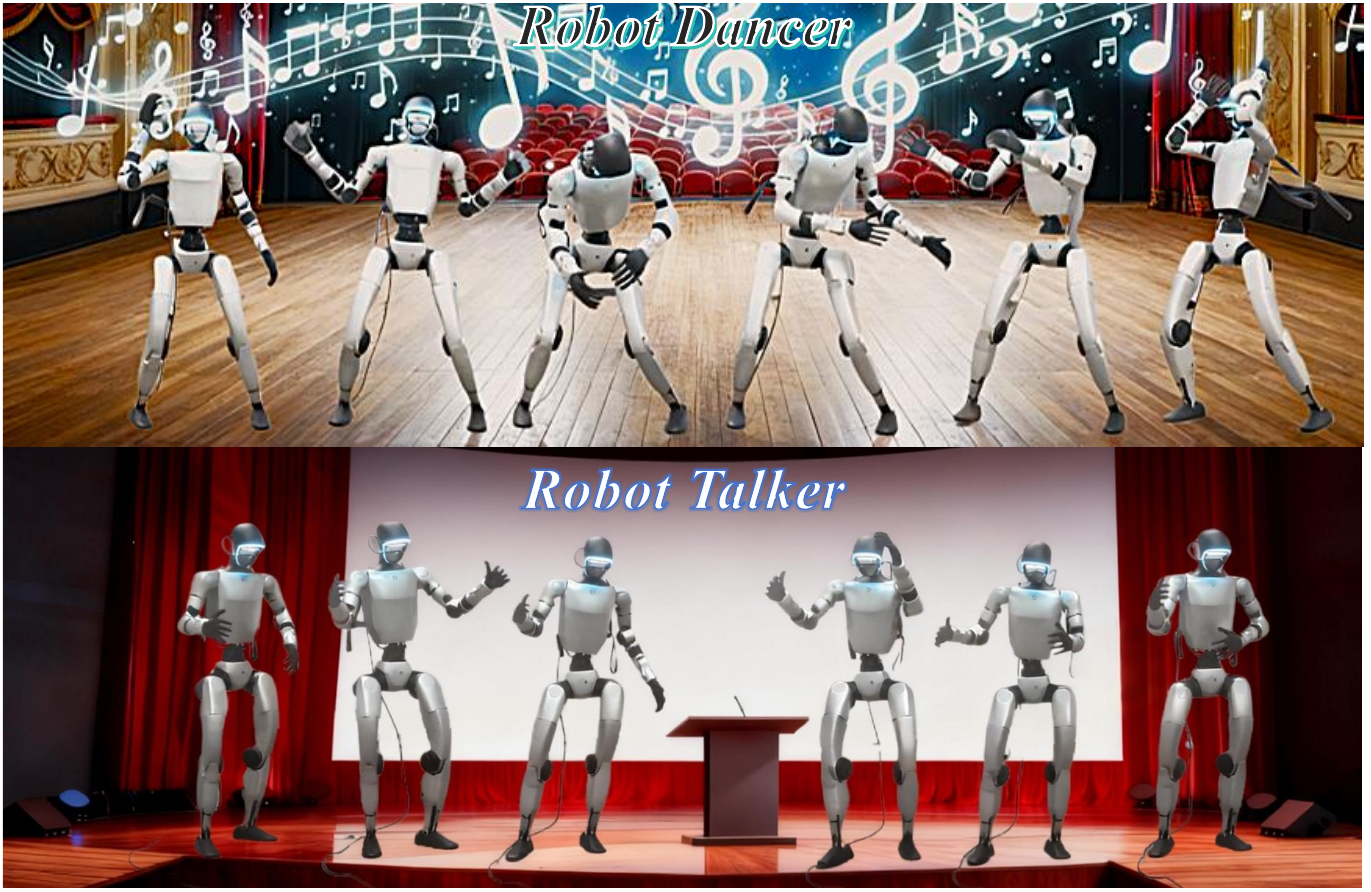}
    \vspace{-3mm}
    \captionof{figure}{\logo \textbf{RoboPerform} makes humanoid perform as dancer and talker， which utilizes audio as signal to control humanoid locomotion, enabling poolicy to generate rhythm-aligned co-speech gestures and dance movements via input speech or music.}
    \label{fig: teaser}
\end{center}%
}]

\begin{abstract}
Humans intuitively move to sound, but current humanoid robots lack expressive improvisational capabilities, confined to predefined motions or sparse commands. Generating motion from audio and then retargeting it to robots relies on explicit motion reconstruction, leading to cascaded errors, high latency, and disjointed acoustic-actuation mapping. We propose RoboPerform, the first unified audio-to-locomotion framework that can directly generate music-driven dance and speech-driven co-speech gestures from audio. Guided by the core principle of ``motion = content + style'', the framework treats audio as implicit style signals and eliminates the need for explicit motion reconstruction. RoboPerform integrates a ResMoE teacher policy for adapting to diverse motion patterns and a diffusion-based student policy for audio style injection. This retargeting-free design ensures low latency and high fidelity. Experimental validation shows that RoboPerform achieves promising results in physical plausibility and audio alignment, successfully transforming robots into responsive performers capable of reacting to audio.

\end{abstract}    
\section{Introduction}
\label{sec:intro}
Humans move to sound. A drumbeat invites a step; a rising melody prompts a leap; spoken emphasis naturally evokes a gesture. 
These responses are not mere kinematic mimicry but arise from an intrinsic understanding of rhythm, phrasing, and intent, which is a process where perception precedes imitation. 
In contrast, most humanoid locomotion systems today are either constrained to mimic pre-defined motion clips~\cite{chen2025gmt, ji2024exbody2, he2024omnih2o, xie2025kungfubot, han2025kungfubot2, he2025hover, mao2025universal, zhou2025roborefer, zhou2025robotracer} or to follow sparse language commands~\cite{li2025language, yue2025rl, shao2025langwbc}.
While effective for simple scripting, these interfaces lack the capacity for expressive, context-sensitive control, and they bypass a crucial question for performative robots: Do you have freestyle?

We argue that humanoid locomotion is fundamentally a generative problem: given a conditioning signal, synthesize physically plausible, stylistically aligned, and semantically grounded motion. This view invites richer modalities beyond text and motion capture, particularly audio, which is dense in temporal structure yet compact to transmit. Music encodes beat, tempo, and timbre that shape movement style; speech carries prosody, emphasis, and discourse rhythm that cue co-speech gestures. Treating audio as a first-class control signal transforms the robot from a replica to a performer: from mechanically replaying dance poses to improvising to the soundtrack; from reading a script to speaking with embodied gestures.

However, dominant pipelines are ill-suited for audio-conditioned control. Explicitly generating human motion via audio-driven motion generators~\cite{bian2025motioncraft, li2025omnimotion, liu2024emage}, followed by retargeting and tracking to the robot via a controller, inherently introduces three systemic issues: 
(1) cascaded error accumulation across decoding, retargeting, and tracking, which degrades both expressive fidelity and physical consistency; 
(2) significant inference latency induced by sequential multi-stage processing, which hinders practical deployment and rapid iteration;
(3) loose coupling between high-level acoustic cues and low-level joint actuation, each module is optimized in isolation, failing to preserve fine-grained expressions such as style, timing, and dynamics. 
Building on this observation, a more direct and natural insight emerges: bypass explicit motion reconstruction, directly encode raw audio, and treat stylistic elements (e.g., beats, prosody, and energy envelopes) as implicit control signals to modulate and refine humanoid locomotion.

Our key insight is simple: \textit{\textbf{motion = content + style}}. 
Building on latent motion representations~\cite{li2025language}, we define \textbf{\textit{content}} as a high-level motion latent which is encoded from a text command (e.g., ``a person is dancing'') via a text-to-motion model to specify the core task.
We treat \textbf{\textit{style}} as the audio signal (e.g., music beats or speech prosody), which dictates how that task is performed. 
We introduce \textit{\textbf{RoboPerform}}, a teacher-student framework designed to realize this decomposition. 
The teacher policy utilizes a $\Delta$MoE, a residual mixture-of-experts architecture, where its experts specialize in diverse motion regimes and complement one another. 
This knowledge is then distilled into the student policy, a diffusion-based generator. 
This student policy explicitly decomposes the generation: it is conditioned on the content latent to preserve the core task, while simultaneously injecting the audio-driven style latents.
This design achieves our goal, enabling the robot to perform the core task while precisely aligning its movements with acoustic details, such as synchronized steps to the beat and nuanced gestures aligned with prosody.

Concretely, RoboPerform guides its diffusion policy using two distinct sets of latents: high-level content latents that define the core task, and temporally-aligned style latents that encode kinematic and prosodic details. These combined latents serve as expressive anchors that guide the student policy to denoise executable actions on the humanoid. This retargeting-free, latent-driven design improves overall inference efficiency, enhances motion fidelity, and ensures fine-grained temporal alignment via the motion latent space. It scales across behaviors, from rhythm- and genre-conditioned freestyle dance to presenter-style co-speech gestures that improve clarity and engagement.


Extensive experiments validate the effectiveness and practicality of RoboPerform across both music-to-dance and speech-to-gesture. RoboPerform delivers temporally aligned, physically plausible motion with smoother style control and significantly higher inference efficiency than retargeting-based pipelines. We further demonstrate its capabilities, enabling humanoids to perform freestyle dance to music and function as hosts, which are presented in Figure \ref{fig: teaser}. In short, \textit{\textbf{RoboPerform reframes humanoid control around audio, moving from motion replay to responsive performance.}}

Our contributions can be summarized as follows:
\begin{itemize}
    \item To our knowledge, RoboPerform is the first framework to utilize audio as an implicit control modality for unified humanoid locomotion and gestural expression, bridging what is heard with how a humanoid moves.
    
    \item We propose $\Delta$MoE in the teacher policy specializes in diverse motion regimes via a mixture-of-experts design, while the student policy decomposes motion into content and style to inject audio-driven style signals into a diffusion-based generator, preserving timing fidelity and reducing end-to-end latency.
    
    \item We validate RoboPerform through extensive experiments across music-to-dance and speech-to-gesture tasks, demonstrating physically plausible, stylistically aligned, and real-time synchronized motion, enabling freestyle performance and embodied speech gestures.

\end{itemize}

\section{Related Work}
\subsection{Humanoid Whole-body Control}
Traditional model-based whole-body control methods achieve precise task execution via accurate dynamics models~\cite{geyer2003positive, sreenath2011compliant}, but suffer from intricate modeling and limited generalization across skills or unmodeled dynamics. Learning-based paradigms rely on manually designed task-specific rewards, succeeding in locomotion~\cite{wang2025beamdojo}, jumping~\cite{peng2021amp}, and fall recovery~\cite{li2023robust, huang2025learning, he2025learning} while requiring elaborate reward engineering and struggling to generate human-like motions. Some studies decompose control into independent policies~\cite{zhang2025falcon, li2025hold}, compromising inter-body coordination, while others use hierarchical frameworks for tasks like table tennis~\cite{su2025hitter}. Whole-body motion tracking offers a paradigm shift~\cite{han2025kungfubot2}: it takes human motion as reference, formulating a unified control goal that obviates task-specific reward design and inherently fosters human-like coordination across diverse skills.

\subsection{Humanoid Motion Tracking}
Humanoid motion tracking learns lifelike behaviors from human motion data. DeepMimic~\cite{peng2018deepmimic} pioneers a phase-based framework with random initialization and early termination for single-motion imitation. ASAP~\cite{he2025asap} addresses the sim-to-real gap via a multi-stage pipeline with a delta-action model for dynamic skills. HuB~\cite{zhang2025hub} and KungfuBot~\cite{xie2025kungfubot} use elaborate processing to accurately imitate highly dynamic single motions.

For unified multi-motion policies, OmniH2O~\cite{he2024omnih2o} introduces a universal controller inspiring subsequent works. ExBody2~\cite{ji2024exbody2} enhances expressiveness via target decomposition and filtering. TWIST~\cite{ze2025twist} and CLONE~\cite{li2025clone} achieve high-quality tracking but are tailored to teleoperation and low-dynamic motions. BumbleBee~\cite{wang2025experts} uses motion clustering, expert policy training, and distillation. GMT~\cite{chen2025gmt} enables robust dynamic motion tracking by prioritizing root velocity and pose over global position. UniTracker~\cite{yin2025unitracker} supports dynamic tracking but lacks stability in long sequences due to global target dependence. BeyondMimic~\cite{liao2025beyondmimic} achieves high-fidelity single-motion tracking via specialized objectives and system identification, further using a distilled diffusion policy for task control. Kungfubot2 proposes an orthogonal MoE for general motion tracking, enabling versatile skill learning. Building on these, we develop a universal policy for audio-driven humanoid action generation, endowing humanoids with the ability to perform.

\subsection{Modality-driven Humanoid Locomotion}
Recent works explore language-guided locomotion. LangWBC~\cite{shao2025langwbc} trains a compact auxiliary network for online motion generation but lacks scalability to complex distributions and unseen instructions. RLPF~\cite{yue2025rl} finetunes an LLM with physical feasibility feedback from a tracking policy to align semantics with kinematics, but risks catastrophic forgetting due to decoder-focused gradient updates. RoboGhost~\cite{li2025language} proposes a latent-driven retargeting-free framework to reduce error accumulation and latency, treating locomotion as a generation task but only using language as input. In this work, we first leverage audio modality as a conditioning signal for humanoid locomotion, achieving "motion synchronized with sound."

\section{Method}

\begin{figure*}[h]
\centering
  \includegraphics[width=2.0\columnwidth, trim={0cm 0cm 0cm 0cm}, clip]{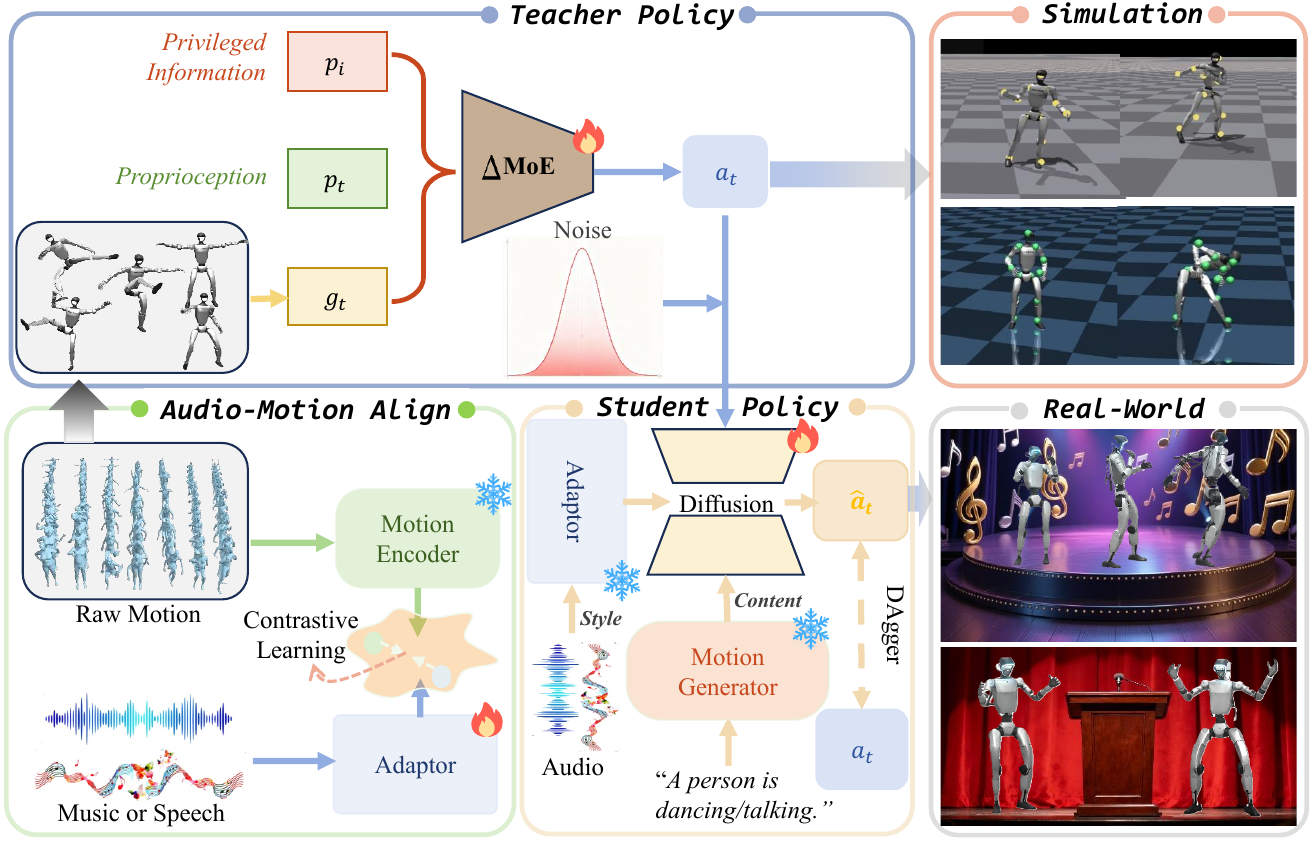}
\caption{Overview of RoboPerform. We propose a two-stage approach: train an adaptor to inject kinematic information into audio modality, then a $\Delta$MoE teacher policy is trained with RL and a diffusion-based student policy is trained to denoise actions conditioned on audio latent. We propose that motion=content+style. Thus, we fix the motion latent as a constant condition and leverage different audio signals as style modulation signals to generate actions adaptive to diverse rhythms.}
\label{fig:framework}
\end{figure*}

\subsection{Overview}
We present a novel audio-driven framework for humanoid motion generation, eliminating error-prone retargeting to enable stylistically aligned, physically plausible actions via fused audio semantics and motion control. 
As shown in Figure~\ref{fig:framework}, its core includes three components: a Delta Mixture of Experts ($\Delta$MoE) teacher policy, an InfoNCE-optimized audio-motion alignment module, and a diffusion-based student policy with content-style disentanglement. It addresses generating expressive motions (e.g., dance and gesture) directly from audio without motion templates or pose estimation.

It begins with audio-motion alignment: an adaptor augmented with temporal attention processes raw audio latents $l_{\text{audio}}$, aligning them with motion latents $l_{\text{motion}}$ via the InfoNCE loss. This design embeds kinematic priors into audio latents, obviating the need for a dedicated audio-to-motion generator and ensuring rhythmic consistency between audio and motion. For robust teacher policy training, we propose $\Delta$MoE, which partitions 3D conditional inputs into nested subspaces $\{S_i\}_{i=1}^4$ for four experts. A gating network dynamically weights experts via residual fusion ($\mathbf{a} = w_1\mathbf{a}_1 + \sum_{i=2}^4 w_i(\mathbf{a}_i - \mathbf{a}_{i-1})$), eliminating redundancy and enhancing expert complementarity. We then distill this oracle policy into a diffusion-based student policy grounded in the "motion=content+style" insight: motion latents from pretrained motion generator guide denoising, while aligned audio latents are injected across diffusion layers to modulate rhythmic expression.

By integrating alignment, specialized teaching, and disentangled diffusion control, our framework achieves direct audio-to-action mapping with low latency and strong generalization. It uniquely enables audio-driven freestyle dance and speech-accompanied gestures, setting a new paradigm for retargeting-free, expressive humanoid control.

\subsection{Delta Mixture of Experts}
\begin{figure}[h]
\centering
  \includegraphics[width=1.0\columnwidth, trim={0cm 0cm 0cm 0cm}, clip]{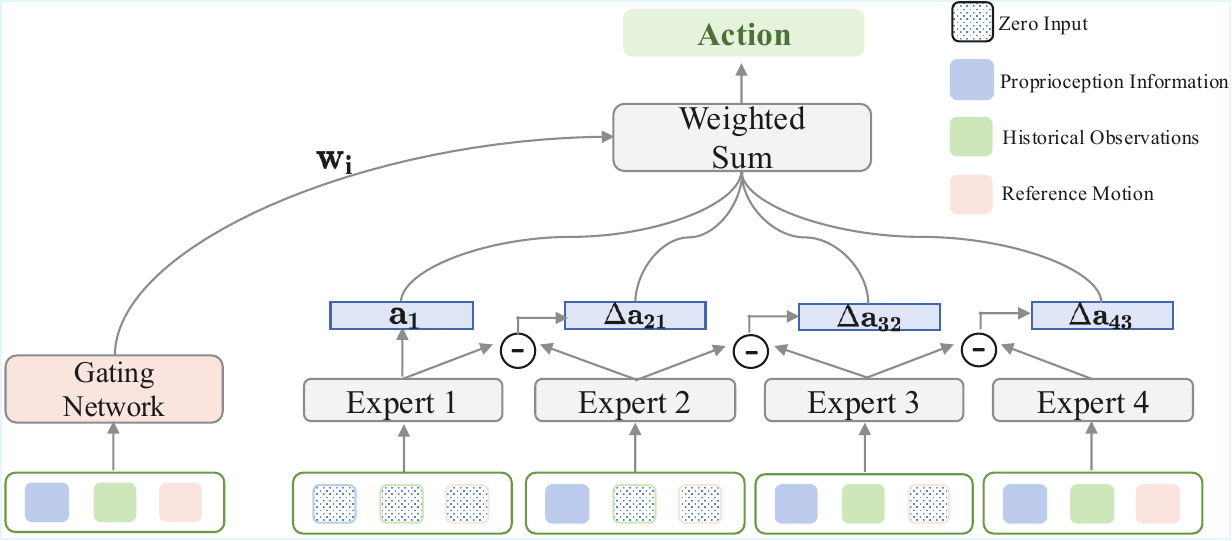}
\caption{Overview of $\Delta$MoE.}
\label{MoE}
\end{figure}

To maximize the diversity and complementarity of knowledge learned by different components, each of which processes a distinct subset of input conditions, we propose $\Delta$MoE as the teacher policy in Figure \ref{MoE}. The core design of $\Delta$MoE hinges on nested conditional subspace partitioning and residual incremental learning, which enforces mutual complementarity among experts while eliminating information redundancy. Fundamentally, $\Delta$MoE can be interpreted as a structured generalization of Classifier-Free Guidance (CFG)~\cite{ho2022classifier} to continuous, multi-dimensional conditional settings, providing a rigorous theoretical foundation for its residual fusion mechanism.

We formalize conditional inputs as a 3D vector $\mathbf{c} = [c_1,c_2,c_3]^T \in \mathbb{R}^3$. In standard CFG, models are trained in both conditional $p(\mathbf{a} \mid \mathbf{c})$ and unconditional $p(\mathbf{a})$ forms, with inference leveraging interpolated fusion:
\[
\mathbf{a} \propto \frac{p(\mathbf{a} \mid \mathbf{c})^\gamma}{p(\mathbf{a})^{1-\gamma}}
\]
In the log-space, this translates to an additive update that balances conditional alignment and unconditional diversity:
\[
\nabla_{\mathbf{a}} \log p(\mathbf{a} \mid \mathbf{c}) + (\gamma-1)\nabla_{\mathbf{a}} \log \frac{p(\mathbf{a} \mid \mathbf{c})}{p(\mathbf{a})}
\]
$\Delta$MoE extends this core insight to a nested hierarchy of partial conditions, defining a filtration of conditional subspaces:
\[
\{0\} = S_1 \subset S_2 \subset S_3 \subset S_4 = \mathbb{R}^3
\]

$\Delta$MoE employs 4 experts $\{e_i\}_{i=1}^4$, where each expert $e_i$ models a policy $\pi_i(\mathbf{a} \mid \mathbf{c}_{S_i})$ that depends solely on the subspace $S_i$: $e_1$ takes $S_1=\{0\}$ as input (modeling the unconditional prior $p(\mathbf{a})$), $e_2$ conditions on $S_2=\{c_1,0,0\}$, $e_3$ uses $S_3=\{c_1,c_2,0\}$ as input, $e_4$ conditions on $S_4=\{c_1,c_2,c_3\}$, modeling the full conditional $p(\mathbf{a} \mid \mathbf{c})$.

A gating network processes $\mathbf{c}$ to output normalized weights $\mathbf{w} = [w_1,...,w_4]^T$ ($\sum w_k = 1$). Residual fusion yields the final action:
\[
\mathbf{a} = w_1\mathbf{a}_1 + \sum_{i=2}^4 w_i(\mathbf{a}_i - \mathbf{a}_{i-1})
\]
where $\mathbf{a}_i$ is $e_i$'s output, and $\Delta\mathbf{a}_i = \mathbf{a}_i - \mathbf{a}_{i-1}$ ($\mathbf{a}_0 = \mathbf{0}$) denotes the marginal contribution of introducing the $i$-th conditional dimension. This formulation is equivalent to a weighted sum of conditional increments:
\[
\mathbf{a} = \sum_{i=1}^4 w_i \Delta\mathbf{a}_i, \quad \text{where } \Delta\mathbf{a}_i = \mathbb{E}[\mathbf{a} \mid \mathbf{c}_{S_i}] - \mathbb{E}[\mathbf{a} \mid \mathbf{c}_{S_{i-1}}]
\]

Each $\Delta\mathbf{a}_i$ directly analogizes to the guidance term in CFG, quantifying the ``information gain'' from adding the $i$-th conditional dimension, just as CFG's residual term disentangles conditional and unconditional signals, $\Delta\mathbf{a}_i$ ensures non-overlapping contributions across experts. This disentanglement eliminates information redundancy while enforcing mutual complementarity: experts do not compete for shared signals but instead specialize in distinct conditional increments.

We adopt $\Delta$MoE as our oracle policy, which takes both robot state observations and reference motion as input, and outputs the final action $\mathbf{a}_t$ optimized with several rewards. By generalizing CFG's residual contrast to hierarchical conditional subspaces, $\Delta$MoE achieves both precise conditional alignment via structured incremental learning and robust generalization via complementary expert knowledge.

\subsection{Audio-Motion Alignment}


To directly guide action generation by conditioning the policy on the audio latent, thereby circumventing the need to train a dedicated audio-to-motion generator, we endow the audio latent with kinematic information. Specifically, we train an audio adaptor to align the audio latent $l_{\text{audio}}$ with the motion latent $l_{\text{motion}}$. The adaptor consists of a 6-layer Transformer~\cite{vaswani2017attention} that processes the audio latent, augmented with temporal attention to capture rhythmic structures inherent in the audio. The motion latent is extracted from our pretrained VAE, and the entire alignment process is optimized using the InfoNCE loss~\cite{oord2018representation}, effectively embedding kinematic priors into the audio latent through the adaptor.

Formally, given a batch of $N$ paired audio-motion latents $\{(l_{\text{audio}}^{(i)}, l_{\text{motion}}^{(i)})\}_{i=1}^N$, we treat each pair $(i,i)$ as a positive sample and all other $(i,j)$ with $j \neq i$ as negative samples. Let $\text{sim}(u, v) = \frac{u^\top v}{\tau}$ denote the scaled cosine similarity between two normalized latent vectors, where $\tau > 0$ is a temperature hyperparameter. The InfoNCE loss is then defined as:
\begin{equation}
    \mathcal{L}_{\text{InfoNCE}} = -\frac{1}{N} \sum_{i=1}^N \log \frac{\exp\left(\text{sim}(l_{\text{audio}}^{(i)}, l_{\text{motion}}^{(i)})\right)}{\sum_{j=1}^N \exp\left(\text{sim}(l_{\text{audio}}^{(i)}, l_{\text{motion}}^{(j)})\right)}.
\end{equation}
This objective encourages the adaptor to map audio latents closer to their corresponding motion latents in the embedding space while pushing them away from unrelated ones.

\subsection{Audio-conditioned Policy Distill}
We posit that \textbf{motion=content+style}. In the context of dance or gesture, the audio serves primarily as a style cue, modulating the underlying motion content in accordance with its rhythmic and temporal structure. To instantiate this disentanglement, we first encode high-level semantic descriptions, e.g., “The person is dancing to the music” or “The person is giving a speech” into a motion latent using a pretrained motion generator. During training, all motions share the same motion latent, which provides the content of the generated action. This motion latent is then employed as the primary conditioning signal to guide the denoising process in the diffusion model.  

Subsequently, the aligned audio latent is injected as an external style control signal into the diffusion backbone at multiple layers, which can be formulated as:

\begin{equation}
    \mathbf{o}_i = \text{Layer}_i\big(\mathbf{o}_{i-1},\, l_{\text{motion}}\big) + \alpha l_{\text{audio}},
\end{equation}
where $\mathbf{o}_i$ denotes the output of layer $i$. This progressive injection steers the denoising trajectory toward rhythmically stylized motion, effectively modulating the base motion content in an audio-aware manner. 

Follwing a DAgger-like approach~\cite{ross2011reduction},  we roll out the student policy in simulation and query the teacher for optimal actions $\hat{\mathbf{a}}$ at visited states. We employ a diffusion model as the student policy to perform action denoising. The forward process progressively corrupts the clean action $\mathbf{a}$ by adding Gaussian noise over $T$ timesteps, yielding noisy samples $\mathbf{x}_t = \sqrt{\bar{\alpha}_t}\, \mathbf{a} + \sqrt{1 - \bar{\alpha}_t}\, \boldsymbol{\epsilon}$, where $\boldsymbol{\epsilon} \sim \mathcal{N}(\mathbf{0}, \mathbf{I})$ and $\bar{\alpha}_t = \prod_{s=1}^t \alpha_s$ denotes the cumulative signal-to-noise ratio at timestep $t$. For tractability, we adopt an $\mathbf{x}_0$-prediction parameterization, where the student policy $\epsilon_\theta(\mathbf{x}_t, t)$ is trained to predict the original clean action $\mathbf{a}$. Specifically, we define the reconstructed action as
$\hat{\mathbf{a}}_t = \frac{\mathbf{x}_t - \sqrt{1 - \bar{\alpha}_t}\, \epsilon_\theta(\mathbf{x}_t, t)}{\sqrt{\bar{\alpha}_t}}$ and supervise the model by minimizing the mean squared error loss $\mathcal{L} = \left\| \mathbf{a} - \hat{\mathbf{a}}_t \right\|_2^2$.
\section{Experiments}
We evaluate RoboPerform on two tasks, including music-driven and speech-driven humanoid control, and rigorously assess whether humanoid locomotion can be effectively generated from audio alone. Specifically, the input audio is first encoded and then processed by a pretrained adaptor to produce a representation aligned with the motion latent space, which is subsequently fed into a policy network to generate executable actions. In our experiments, both the teacher and student policies are trained in the IsaacGym simulation environment, and the student policy is directly deployed on the Unitree G1 humanoid robot for real-world validation.
\subsection{Experimental Setups}
\paragraph{Dataset}
We train our model on FineDance~\cite{li2023finedance} and BEAT2~\cite{liu2024emage} datasets. BEAT2 has 76 hours of data from 30 speakers, standardized into a mesh representation with paired audio. FineDance is a fine-grained 3D full-body dataset, which has 7.7 hours of dance motion. It provides the SMPL-H~\cite{pavlakos2019expressive} format motion data and music feature extracted by librosa. All motions are sampled at 30FPS. Due to the excessive length of the original data, we segment each motion sequence and its corresponding audio into 10-second clips for both training and evaluation.

\paragraph{Metrics}
We adopt two categories of evaluation metrics: audio-motion retrieval and motion tracking. For audio-motion retrieval, we only report retrieval precision R@{1,2,3} to evaluate the ability of audio adaptor. For motion tracking, evaluated in physics simulators aligning with prior works~\cite{he2024omnih2o}, we use success rate as the core indicator, supplemented by mean per-joint position error ($E_{\text{MPJPE}}$) and mean per-keypoint position error ($E_{\text{MPKPE}}$). Detailed metric definitions are provided in the Appendix.

\paragraph{Implementation Details}
In $\Delta$MoE, we employ 4 MLP-based experts together with an MLP gating network that assigns a weight to each expert. Since the FineDance dataset provides pre-encoded music features, we do not train a music encoder; however, for speech from the BEAT2 dataset, we adopt the temporal convolutional network from EMAGE~\cite{liu2024emage} to learn speech representations. We train a 9-layer, 4-head transformer as the motion VAE and a 6-layer, 4-head transformer as the music adaptor, where temporal attention is explicitly incorporated to capture temporal dynamics. During DAgger training, we utilize a 4-layer MLP as the backbone of the diffusion model, with conditioning injected via AdaLN~\cite{huang2017arbitrary}. At inference, we employ a two-step DDIM sampling~\cite{song2020denoising} schedule to ensure real-time performance during deployment. Further details regarding policy training can be found in the Appendix.

\begin{figure}[t]
\centering
  \includegraphics[width=0.9\columnwidth]{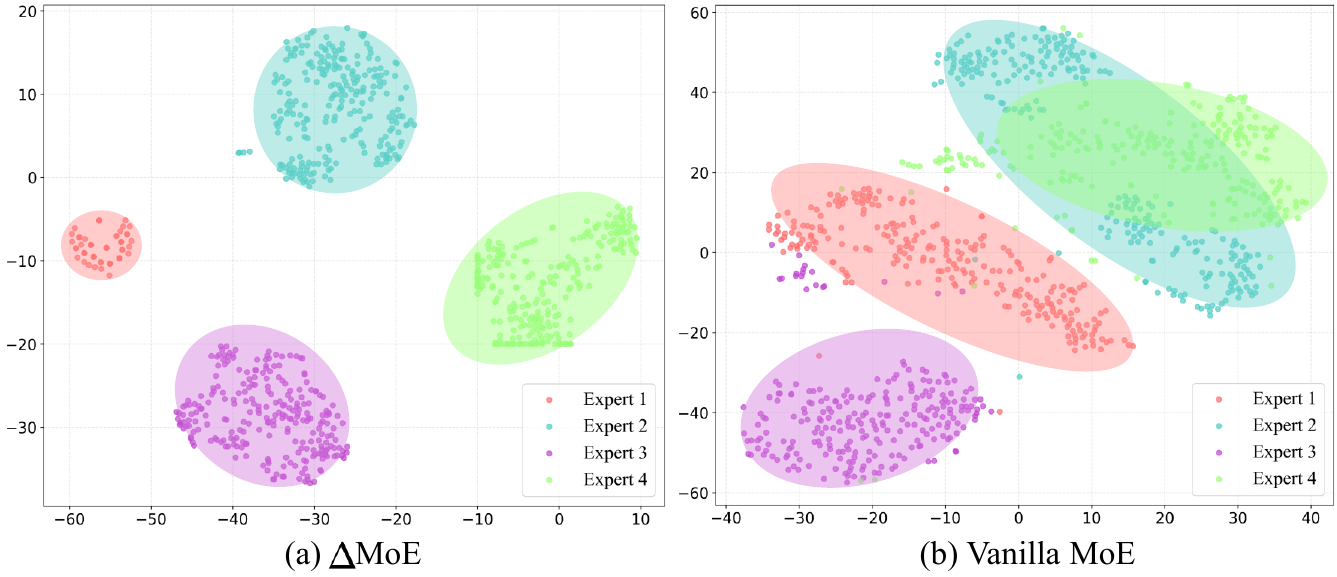}
\caption{T-SNE visualization results of each component for $\Delta$MoE and vanilla MoE.}
\label{fig:tnse}
\end{figure}

\begin{table}[t]\large
\centering
\small
\setlength{\tabcolsep}{2.5pt}
\begin{tabular}{l|cccc}
\toprule
{Method} & R@1 $\uparrow$ & R@2 $\uparrow$ & R@3 $\uparrow$ & MM-Dist $\downarrow$ \\
\midrule
Music-Motion &  66.7 & 78.8 & 83.5 & 1.154\\
Speech-Motion &  64.6 & 76.5 & 82.1 & 1.232\\
\bottomrule
\end{tabular}
\caption{Audio-motion alignment performance on the BEAT2 and FineDance test sets.}
\label{retrieval}
\end{table}

\subsection{Evaluation of Audio-Motion Retrieval}
To evaluate the alignment capability of the audio adaptor, we conduct alignment evaluation on the test sets of FineDance and BEAT2, specifically assessing whether the model can accurately align a given audio segment to the motion latent space to retrieve the corresponding motion latent. The results are shown in Table \ref{retrieval}.

\begin{figure}[t]
\centering
  \includegraphics[width=0.85\columnwidth]{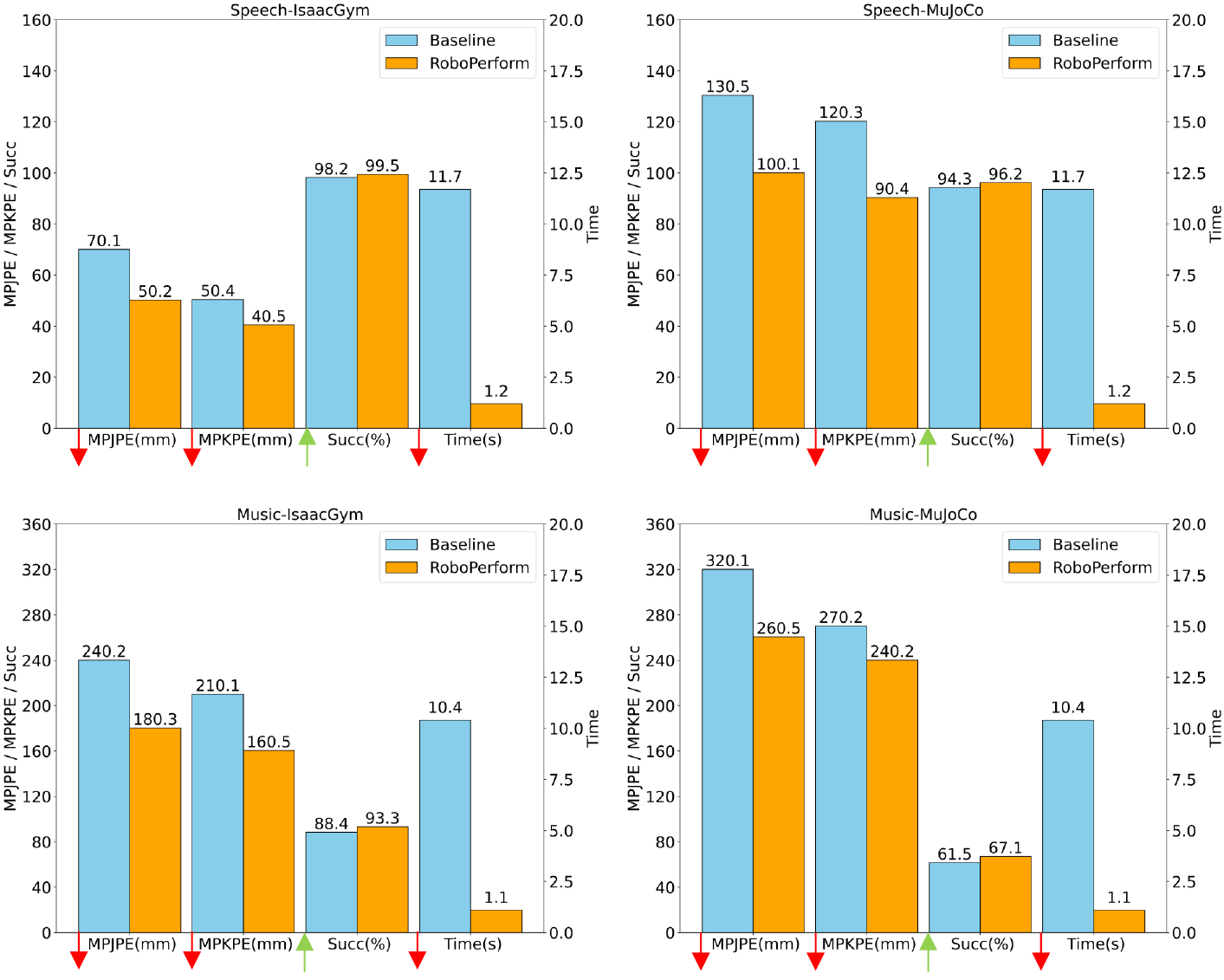}
\caption{Ablation study on tracking performance of music-to-dance and speech-to-gesture tasks in IsaacGym and MuJoCo. The baseline uses pretrained motion generators for each task to generate motions, which drive the student policy for action generation.}
\label{fig:ab1}
\end{figure}

\subsection{Evaluation of Motion Tracking}
To further validate the motion tracking performance of our policy, we evaluate it on two audio-driven locomotion tasks: speech-to-locomotion and music-to-locomotion, reporting task success rate, $E_{\text{mpjpe}}$, $E_{\text{mpkpe}}$ in IsaacGym and MuJoCo. The pipeline operates as follows: (i) the input audio is encoded and processed by our trained adaptor to inject kinematic priors into the audio features; (ii) the resulting motion-aligned latent representation conditions the student policy to generate physically executable actions. As shown in Table~\ref{tab:tracking_results}, our method achieves high task success rates on both the FineDance and BEAT2 datasets, along with low joint and keypoint errors, indicating strong alignment between audio semantics and feasible locomotion trajectories. The baseline employs pretrained models EMAGE~\cite{liu2024emage} and FineNet~\cite{li2023finedance} to first generate a deterministic motion, which is then retargeted to G1 and executed by an explicit motion-driven policy based MLP.

\begin{figure*}[ht]
\centering
  \includegraphics[width=1.75\columnwidth]{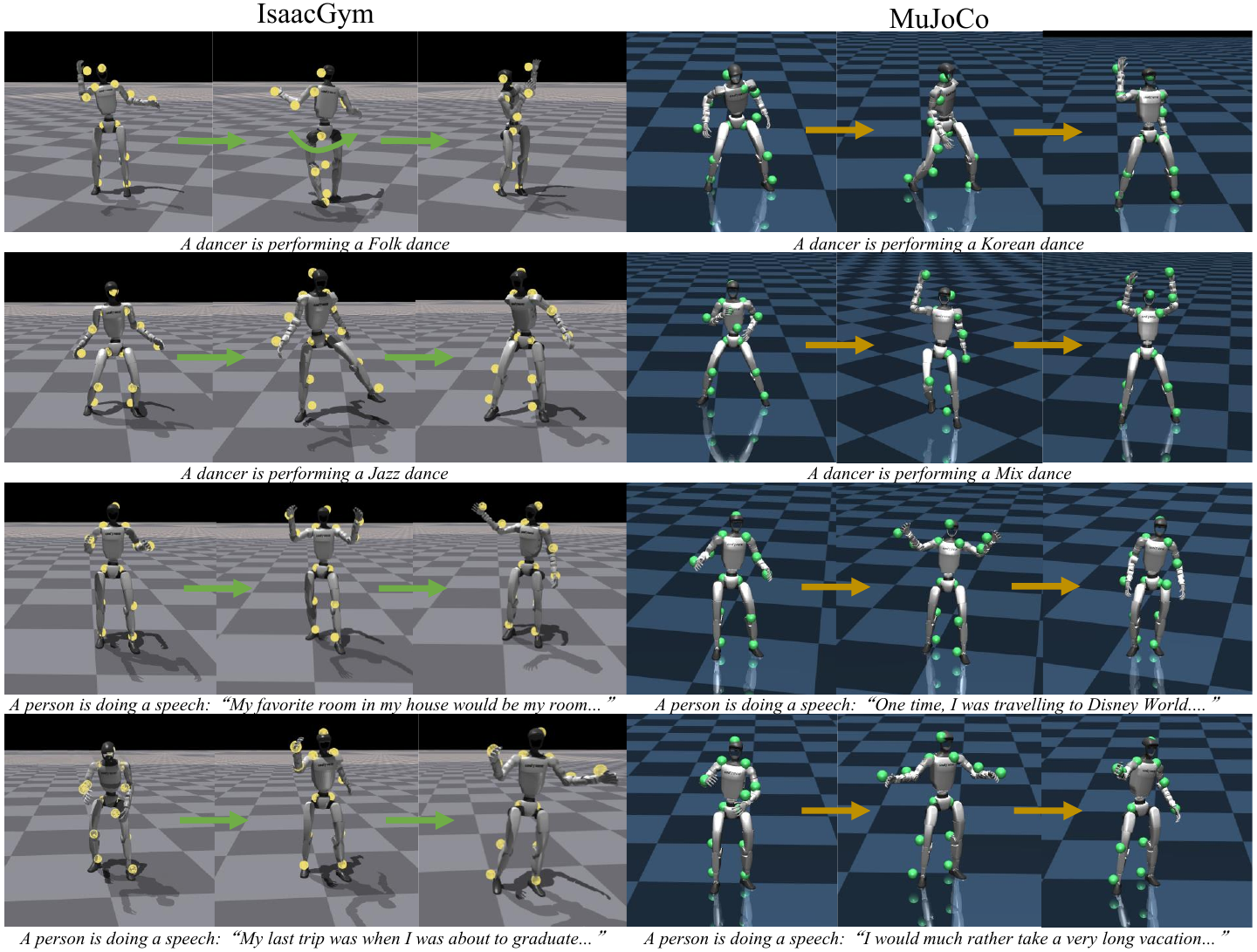}
\caption{Qualitative results in the IsaacGym and MuJoCo. The upper half presents the tracking performance of music-to-locomotion, and the lower half presents that of speech-to-locomotion.}
\label{fig:qualitative}
\end{figure*}

\begin{table}[t]\large
\centering
\footnotesize 
\setlength{\tabcolsep}{2pt}
\begin{tabular}{lcccccc}
\toprule
\multirow{2}{*}{Method} & \multicolumn{3}{c}{IsaacGym} & \multicolumn{3}{c}{MuJoCo} \\
\cmidrule(lr){2-4} \cmidrule(lr){5-7}
& Succ $\uparrow$ & $E_{mpjpe}$ $\downarrow$ & $E_{mpkpe}$ $\downarrow$ & Succ $\uparrow$ & $E_{mpjpe}$ $\downarrow$ & $E_{mpkpe}$ $\downarrow$ \\
\midrule
\multicolumn{7}{c}{BEAT2} \\
\midrule
Baseline     & \cellcolor{color1}0.98 &\cellcolor{color1}0.07 & \cellcolor{color1}0.05 &\cellcolor{color1} 0.94 &\cellcolor{color1} 0.13 & \cellcolor{color1}0.12 \\
Ours          & \cellcolor{color2}0.99 &\cellcolor{color2} 0.05 & \cellcolor{color2}0.04 &\cellcolor{color2} 0.96 &\cellcolor{color2} 0.10 &\cellcolor{color2} 0.09 \\
\midrule
\midrule
\multicolumn{7}{c}{FineDance} \\
\midrule
Baseline    & \cellcolor{color1}0.88 &\cellcolor{color1}0.24 & \cellcolor{color1}0.21 &\cellcolor{color1} 0.61 &\cellcolor{color1} 0.32 & \cellcolor{color1}0.27\\
Ours         &\cellcolor{color2} 0.93 & \cellcolor{color2}0.18 & \cellcolor{color2}0.16 & \cellcolor{color2}0.67 & \cellcolor{color2}0.26 & \cellcolor{color2}0.24 \\
\bottomrule
\end{tabular}
\caption{Motion tracking performance comparison in simulation on the BEAT2 and FineDance test sets.}
\label{tab:tracking_results}
\end{table}

\subsection{Qualitative Results}
We conduct a qualitative assessment of the motion tracking policy across three deployment settings: simulation (IsaacGym), cross-simulator transfer (MuJoCo), and real-world execution on the Unitree G1 humanoid robot. Figure~\ref{fig:qualitative} presents representative tracking sequences, highlighting the policy’s ability to adhere to audio rhythm, maintain balance during dynamic transitions, and generalize across diverse physics engines and hardware platforms. Additional qualitative results in simulation and real-world settings are provided in Appendix.

\begin{table}[t]\large
\centering
\footnotesize   
\setlength{\tabcolsep}{1pt}
\begin{tabular}{lcccccc}
\toprule
\multirow{2}{*}{Method} & \multicolumn{3}{c}{IsaacGym} & \multicolumn{3}{c}{MuJoCo} \\
\cmidrule(lr){2-4} \cmidrule(lr){5-7}
& Succ $\uparrow$ & $E_{mpjpe}$ $\downarrow$ & $E_{mpkpe}$ $\downarrow$ & Succ $\uparrow$ & $E_{mpjpe}$ $\downarrow$ & $E_{mpkpe}$ $\downarrow$ \\
\midrule
\multicolumn{7}{c}{BEAT2} \\
\midrule
Vanilla MoE    & \cellcolor{color1}0.97 &\cellcolor{color1}0.14 & \cellcolor{color1}0.1 &\cellcolor{color1} 0.94&\cellcolor{color1} 0.16 & \cellcolor{color1}0.14 \\
$\Delta$MoE         & \cellcolor{color2}0.99 &\cellcolor{color2} 0.05 & \cellcolor{color2}0.04 &\cellcolor{color2} 0.96 &\cellcolor{color2} 0.10 &\cellcolor{color2} 0.09 \\
\midrule
\midrule
\multicolumn{7}{c}{FineDance} \\
\midrule
Vanilla MoE    & \cellcolor{color1}0.89 &\cellcolor{color1}0.24 & \cellcolor{color1}0.22 &\cellcolor{color1} 0.61 &\cellcolor{color1} 0.29 & \cellcolor{color1}0.26\\
$\Delta$MoE         &\cellcolor{color2} 0.93 & \cellcolor{color2}0.18 & \cellcolor{color2}0.16 & \cellcolor{color2}0.67 & \cellcolor{color2}0.26 & \cellcolor{color2}0.24 \\
\bottomrule
\end{tabular}
\caption{Ablation study on vanilla MoE and $\Delta$MoE across both BEAT2 and FineDance datasets.}
\label{tab:ab1}
\end{table}

\subsection{Ablation Studies}
To systematically validate the effectiveness of the proposed method, we present a set of ablation studies in this section, covering four key aspects: (1) the efficacy of $\Delta$MoE, (2) a comparison between pose-driven and audio-driven locomotion, (3) the necessity of semantic content for locomotion generation, and (4) the necessity of audio adaptor. More ablation studies can be seen in the Appendix.
\paragraph{Audio-driven Vs Pose-driven}
To compare against explicit pose-driven approaches, we generate motions using EMAGE and FineNet and deploy the resulting explicit motion sequences for execution. The time cost reports the full inference latency, baseline encompasses both motion generation and retargeting; specifically, we employ a 1000-iteration PBHC retargeting~\cite{xie2025kungfubot} procedure. As shown in Figure~\ref{fig:ab1}, such explicit action generation not only incurs additional computational overhead but also degrades task success rates and introduces extra tracking error.

\begin{table}[t]\large
\centering
\footnotesize   
\setlength{\tabcolsep}{1pt}
\begin{tabular}{lcccccc}
\toprule
\multirow{2}{*}{Method} & \multicolumn{3}{c}{IsaacGym} & \multicolumn{3}{c}{MuJoCo} \\
\cmidrule(lr){2-4} \cmidrule(lr){5-7}
& Succ $\uparrow$ & $E_{mpjpe}$ $\downarrow$ & $E_{mpkpe}$ $\downarrow$ & Succ $\uparrow$ & $E_{mpjpe}$ $\downarrow$ & $E_{mpkpe}$ $\downarrow$ \\
\midrule
\multicolumn{7}{c}{BEAT2} \\
\midrule
- Content     & \cellcolor{color1}0.96 &\cellcolor{color1}0.11 & \cellcolor{color1}0.09 &\cellcolor{color1} 0.91 &\cellcolor{color1} 0.12 & \cellcolor{color1}0.10 \\
+ Content          & \cellcolor{color2}0.99 &\cellcolor{color2} 0.05 & \cellcolor{color2}0.04 &\cellcolor{color2} 0.96 &\cellcolor{color2} 0.10 &\cellcolor{color2} 0.09 \\
\midrule
\midrule
\multicolumn{7}{c}{FineDance} \\
\midrule
- Content    & \cellcolor{color1}0.91 &\cellcolor{color1}0.20 & \cellcolor{color1}0.17 &\cellcolor{color1} 0.66 &\cellcolor{color2} 0.25 & \cellcolor{color2}0.24\\
+ Content         &\cellcolor{color2} 0.93 & \cellcolor{color2}0.18 & \cellcolor{color2}0.16 & \cellcolor{color2}0.67 & \cellcolor{color1}0.26 & \cellcolor{color2}0.24 \\
\bottomrule
\end{tabular}
\caption{Ablation study on whether to incorporate content information. Herein, the content for both tasks is fixed, with the same content latent used in each inference.}
\label{tab:ab2}
\end{table}

\paragraph{$\Delta$MoE Vs Vanilla MoE}
To investigate the performance gain introduced by our $\Delta$ MoE, we conduct an ablation study comparing the tracking performance of vanilla MoE and $\Delta$ MoE. As shown in Table~\ref{tab:ab1}, $\Delta$ MoE yields consistently more accurate tracking.

Additionally, we visualize every component in MoE using t-SNE. For vanilla MoE, each component corresponds to the output of an individual expert. As illustrated in Figure~\ref{fig:tnse} (b), the features learned by each expert exhibit significant overlap, failing to achieve mutually independent information specialization across experts. In contrast, the components of $\Delta$MoE correspond to the differences between experts conditioned on distinct signals (except for the first expert, which is conditioned on a zero vector). As shown in Figure~\ref{fig:tnse} (a), we perform clustering on \(\{\mathbf{a_1}, \mathbf{a_2 - a_1}, \dots, \mathbf{a_4 - a_3}\}\). The results demonstrate that each component is mutually independent, which fully exploits the capacity of individual experts and enhances generalization. This mechanism is analogous to creating a complete painting: starting with a blank canvas, we incrementally add contour and color information, where each stroke introduces non-redundant details until the artwork is fully realized.
\paragraph{With Content Vs Without Content}
We posit that motion can be decomposed into content and style. For expressive motions such as dance and speech gestures, audio serves primarily as a style modulation signal that shapes the temporal structure, such as rhythm and beat patterns, rather than prescribing fine-grained kinematics. Accordingly, we treat the content latent from a pretrained motion generative model as the primary control signal, and progressively inject audio features into the diffusion process to modulate the denoising trajectory. Herein, LaMP-T2M~\cite{li2024lamp} is adopted as the motion generator for both tasks. For the music-to-dance task, the input text is \textit{"The person is dancing to the music"}, while for the speech-to-gesture task, the input text is \textit{"The person is giving a speech"}. This design ensures that the generated actions preserve semantic content while aligning with the temporal dynamics of the input audio. As shown in Table~\ref{tab:ab2}, policies conditioned on the content latent achieve significantly more accurate tracking performance.

\paragraph{With Adaptor Vs Without Adaptor}
To demonstrate that enriching the control signal with kinematic information leads to more accurate and rhythmically coherent action generation, we conduct an ablation study on the use of the audio adaptor. As shown in Table~\ref{tab:ab3}, when the control signal is aligned with the motion latent space and imbued with kinematic cues via the adaptor, it more effectively guides motion synthesis, yielding improved tracking accuracy and stronger rhythmic alignment. Additionally, we report the rhythm hit rate, a metric quantifying the temporal correspondence between generated motions and musical beats.

\begin{table}[t]\large
\centering
\footnotesize   
\setlength{\tabcolsep}{1pt}
\begin{tabular}{lcccccc}
\toprule
\multirow{2}{*}{Method} & \multicolumn{3}{c}{IsaacGym} & \multicolumn{3}{c}{MuJoCo} \\
\cmidrule(lr){2-4} \cmidrule(lr){5-7}
& Succ $\uparrow$ & $E_{mpjpe}$ $\downarrow$ & $E_{mpkpe}$ $\downarrow$ & Succ $\uparrow$ & $E_{mpjpe}$ $\downarrow$ & $E_{mpkpe}$ $\downarrow$ \\
\midrule
\multicolumn{7}{c}{BEAT2} \\
\midrule
- Adaptor     & \cellcolor{color1}0.88 &\cellcolor{color1}0.29 & \cellcolor{color1}0.27 &\cellcolor{color1} 0.83 &\cellcolor{color1} 0.36 & \cellcolor{color1}0.35 \\
+ Adaptor          & \cellcolor{color2}0.99 &\cellcolor{color2} 0.05 & \cellcolor{color2}0.04 &\cellcolor{color2} 0.96 &\cellcolor{color2} 0.10 &\cellcolor{color2} 0.09 \\
\midrule
\midrule
\multicolumn{7}{c}{FineDance} \\
\midrule
- Adaptor    & \cellcolor{color1}0.79 &\cellcolor{color1}0.49 & \cellcolor{color1}0.48 &\cellcolor{color1} 0.51 &\cellcolor{color1} 0.58 & \cellcolor{color1}0.53\\
+ Adaptor         &\cellcolor{color2} 0.93 & \cellcolor{color2}0.18 & \cellcolor{color2}0.16 & \cellcolor{color2}0.67 & \cellcolor{color2}0.26 & \cellcolor{color2}0.24 \\
\bottomrule
\end{tabular}
\caption{Ablation study on whether to use adaptor inject kinematic information into audio modality. It can be observed that adaptor successfully aligns the audio and motion, improving the tracking performance and success rate.}
\label{tab:ab3}
\end{table}
\section{Conclusion}
We present RoboPerform, a retargeting-free audio-to-locomotion framework that unifies music-driven dance and speech-driven co-speech gesture generation for humanoids. By formulating motion = content + style, our approach leverages a pretrained motion latent for semantic grounding and injects rhythm-aware audio features into a diffusion-based policy. Our proposed $\Delta$MoE enhances behavioral diversity, while content-style disentanglement ensures temporally coherent and physically plausible execution. RoboPerform achieves better tracking performance and faster speed during deployment. It reframes humanoid control as an expressive act, answering the question: \textit{\textbf{Yes, humanoids can freestyle.}}
\section{Acknowledgement}
This work was supported by the National Natural Science Foundation of China (62476011), and by the Beijing Natural Science Foundation (L252060).
{
    \small
    \bibliographystyle{ieeenat_fullname}
    \bibliography{main}

@String(TOG= {ACM Trans. Graph.})

@String(AAAI = {AAAI})

@String(TOG   = {ACM TOG})

@article{li2025language,
  title={From language to locomotion: Retargeting-free humanoid control via motion latent guidance},
  author={Li, Zhe and Chi, Cheng and Wei, Yangyang and Zhu, Boan and Peng, Yibo and Huang, Tao and Wang, Pengwei and Wang, Zhongyuan and Zhang, Shanghang and Xu, Chang},
  journal={arXiv preprint arXiv:2510.14952},
  year={2025}
}

@article{vaswani2017attention,
  title={Attention is all you need},
  author={Vaswani, Ashish and Shazeer, Noam and Parmar, Niki and Uszkoreit, Jakob and Jones, Llion and Gomez, Aidan N and Kaiser, {\L}ukasz and Polosukhin, Illia},
  journal={Advances in neural information processing systems},
  volume={30},
  year={2017}
}

@article{oord2018representation,
  title={Representation learning with contrastive predictive coding},
  author={Oord, Aaron van den and Li, Yazhe and Vinyals, Oriol},
  journal={arXiv preprint arXiv:1807.03748},
  year={2018}
}

@inproceedings{ross2011reduction,
  title={A reduction of imitation learning and structured prediction to no-regret online learning},
  author={Ross, St{\'e}phane and Gordon, Geoffrey and Bagnell, Drew},
  booktitle={Proceedings of the fourteenth international conference on artificial intelligence and statistics},
  pages={627--635},
  year={2011},
  organization={JMLR Workshop and Conference Proceedings}
}

@inproceedings{li2023finedance,
  title={FineDance: A Fine-grained Choreography Dataset for 3D Full Body Dance Generation},
  author={Li, Ronghui and Zhao, Junfan and Zhang, Yachao and Su, Mingyang and Ren, Zeping and Zhang, Han and Tang, Yansong and Li, Xiu},
  booktitle={Proceedings of the IEEE/CVF International Conference on Computer Vision},
  pages={10234--10243},
  year={2023}
}

@inproceedings{liu2024emage,
  title={Emage: Towards unified holistic co-speech gesture generation via expressive masked audio gesture modeling},
  author={Liu, Haiyang and Zhu, Zihao and Becherini, Giorgio and Peng, Yichen and Su, Mingyang and Zhou, You and Zhe, Xuefei and Iwamoto, Naoya and Zheng, Bo and Black, Michael J},
  booktitle={Proceedings of the IEEE/CVF Conference on Computer Vision and Pattern Recognition},
  pages={1144--1154},
  year={2024}
}

@inproceedings{pavlakos2019expressive,
  title={Expressive body capture: 3d hands, face, and body from a single image},
  author={Pavlakos, Georgios and Choutas, Vasileios and Ghorbani, Nima and Bolkart, Timo and Osman, Ahmed AA and Tzionas, Dimitrios and Black, Michael J},
  booktitle={Proceedings of the IEEE/CVF conference on computer vision and pattern recognition},
  pages={10975--10985},
  year={2019}
}

@inproceedings{huang2017arbitrary,
  title={Arbitrary style transfer in real-time with adaptive instance normalization},
  author={Huang, Xun and Belongie, Serge},
  booktitle={Proceedings of the IEEE international conference on computer vision},
  pages={1501--1510},
  year={2017}
}

@article{song2020denoising,
  title={Denoising diffusion implicit models},
  author={Song, Jiaming and Meng, Chenlin and Ermon, Stefano},
  journal={arXiv preprint arXiv:2010.02502},
  year={2020}
}

@article{xie2025kungfubot,
  title={KungfuBot: Physics-Based Humanoid Whole-Body Control for Learning Highly-Dynamic Skills},
  author={Xie, Weiji and Han, Jinrui and Zheng, Jiakun and Li, Huanyu and Liu, Xinzhe and Shi, Jiyuan and Zhang, Weinan and Bai, Chenjia and Li, Xuelong},
  journal={arXiv preprint arXiv:2506.12851},
  year={2025}
}

@article{li2024lamp,
  title={Lamp: Language-motion pretraining for motion generation, retrieval, and captioning},
  author={Li, Zhe and Yuan, Weihao and He, Yisheng and Qiu, Lingteng and Zhu, Shenhao and Gu, Xiaodong and Shen, Weichao and Dong, Yuan and Dong, Zilong and Yang, Laurence T},
  journal={arXiv preprint arXiv:2410.07093},
  year={2024}
}

@inproceedings{he2025hover,
  title={Hover: Versatile neural whole-body controller for humanoid robots},
  author={He, Tairan and Xiao, Wenli and Lin, Toru and Luo, Zhengyi and Xu, Zhenjia and Jiang, Zhenyu and Kautz, Jan and Liu, Changliu and Shi, Guanya and Wang, Xiaolong and others},
  booktitle={2025 IEEE International Conference on Robotics and Automation (ICRA)},
  pages={9989--9996},
  year={2025},
  organization={IEEE}
}

@article{he2024omnih2o,
  title={Omnih2o: Universal and dexterous human-to-humanoid whole-body teleoperation and learning},
  author={He, Tairan and Luo, Zhengyi and He, Xialin and Xiao, Wenli and Zhang, Chong and Zhang, Weinan and Kitani, Kris and Liu, Changliu and Shi, Guanya},
  journal={arXiv preprint arXiv:2406.08858},
  year={2024}
}

@article{ji2024exbody2,
  title={Exbody2: Advanced expressive humanoid whole-body control},
  author={Ji, Mazeyu and Peng, Xuanbin and Liu, Fangchen and Li, Jialong and Yang, Ge and Cheng, Xuxin and Wang, Xiaolong},
  journal={arXiv preprint arXiv:2412.13196},
  year={2024}
}

@article{chen2025gmt,
  title={GMT: General Motion Tracking for Humanoid Whole-Body Control},
  author={Chen, Zixuan and Ji, Mazeyu and Cheng, Xuxin and Peng, Xuanbin and Peng, Xue Bin and Wang, Xiaolong},
  journal={arXiv preprint arXiv:2506.14770},
  year={2025}
}

@article{yue2025rl,
  title={RL from Physical Feedback: Aligning Large Motion Models with Humanoid Control},
  author={Yue, Junpeng and Wang, Zepeng and Wang, Yuxuan and Zeng, Weishuai and Wang, Jiangxing and Xu, Xinrun and Zhang, Yu and Zheng, Sipeng and Ding, Ziluo and Lu, Zongqing},
  journal={arXiv preprint arXiv:2506.12769},
  year={2025}
}

@article{shao2025langwbc,
  title={LangWBC: Language-directed Humanoid Whole-Body Control via End-to-end Learning},
  author={Shao, Yiyang and Huang, Xiaoyu and Zhang, Bike and Liao, Qiayuan and Gao, Yuman and Chi, Yufeng and Li, Zhongyu and Shao, Sophia and Sreenath, Koushil},
  journal={arXiv preprint arXiv:2504.21738},
  year={2025}
}

@article{han2025kungfubot2,
  title={KungfuBot2: Learning Versatile Motion Skills for Humanoid Whole-Body Control},
  author={Han, Jinrui and Xie, Weiji and Zheng, Jiakun and Shi, Jiyuan and Zhang, Weinan and Xiao, Ting and Bai, Chenjia},
  journal={arXiv preprint arXiv:2509.16638},
  year={2025}
}

@inproceedings{bian2025motioncraft,
  title={Motioncraft: Crafting whole-body motion with plug-and-play multimodal controls},
  author={Bian, Yuxuan and Zeng, Ailing and Ju, Xuan and Liu, Xian and Zhang, Zhaoyang and Liu, Wei and Xu, Qiang},
  booktitle={Proceedings of the AAAI Conference on Artificial Intelligence},
  volume={39},
  number={2},
  pages={1880--1888},
  year={2025}
}

@article{li2025omnimotion,
  title={OmniMotion: Multimodal Motion Generation with Continuous Masked Autoregression},
  author={Li, Zhe and Yuan, Weihao and Shen, Weichao and Zhu, Siyu and Dong, Zilong and Xu, Chang},
  journal={arXiv preprint arXiv:2510.14954},
  year={2025}
}

@inproceedings{mao2025universal,
  title={Universal humanoid robot pose learning from internet human videos},
  author={Mao, Jiageng and Zhao, Siheng and Song, Siqi and Hong, Chuye and Shi, Tianheng and Ye, Junjie and Zhang, Mingtong and Geng, Haoran and Malik, Jitendra and Guizilini, Vitor and others},
  booktitle={2025 IEEE-RAS 24th International Conference on Humanoid Robots (Humanoids)},
  pages={1--8},
  year={2025},
  organization={IEEE}
}

@article{peng2018deepmimic,
  title={Deepmimic: Example-guided deep reinforcement learning of physics-based character skills},
  author={Peng, Xue Bin and Abbeel, Pieter and Levine, Sergey and Van de Panne, Michiel},
  journal={ACM Transactions On Graphics (TOG)},
  volume={37},
  number={4},
  pages={1--14},
  year={2018},
  publisher={ACM New York, NY, USA}
}

@article{he2025asap,
  title={Asap: Aligning simulation and real-world physics for learning agile humanoid whole-body skills},
  author={He, Tairan and Gao, Jiawei and Xiao, Wenli and Zhang, Yuanhang and Wang, Zi and Wang, Jiashun and Luo, Zhengyi and He, Guanqi and Sobanbab, Nikhil and Pan, Chaoyi and others},
  journal={arXiv preprint arXiv:2502.01143},
  year={2025}
}

@article{zhang2025hub,
  title={HuB: Learning Extreme Humanoid Balance},
  author={Zhang, Tong and Zheng, Boyuan and Nai, Ruiqian and Hu, Yingdong and Wang, Yen-Jen and Chen, Geng and Lin, Fanqi and Li, Jiongye and Hong, Chuye and Sreenath, Koushil and others},
  journal={arXiv preprint arXiv:2505.07294},
  year={2025}
}

@article{ze2025twist,
  title={Twist: Teleoperated whole-body imitation system},
  author={Ze, Yanjie and Chen, Zixuan and Ara{\'u}jo, Joao Pedro and Cao, Zi-ang and Peng, Xue Bin and Wu, Jiajun and Liu, C Karen},
  journal={arXiv preprint arXiv:2505.02833},
  year={2025}
}

@article{li2025clone,
  title={CLONE: Closed-Loop Whole-Body Humanoid Teleoperation for Long-Horizon Tasks},
  author={Li, Yixuan and Lin, Yutang and Cui, Jieming and Liu, Tengyu and Liang, Wei and Zhu, Yixin and Huang, Siyuan},
  journal={arXiv preprint arXiv:2506.08931},
  year={2025}
}

@article{wang2025experts,
  title={From experts to a generalist: Toward general whole-body control for humanoid robots},
  author={Wang, Yuxuan and Yang, Ming and Ding, Ziluo and Zhang, Yu and Zeng, Weishuai and Xu, Xinrun and Jiang, Haobin and Lu, Zongqing},
  journal={arXiv preprint arXiv:2506.12779},
  year={2025}
}

@article{yin2025unitracker,
  title={Unitracker: Learning universal whole-body motion tracker for humanoid robots},
  author={Yin, Kangning and Zeng, Weishuai and Fan, Ke and Dai, Minyue and Wang, Zirui and Zhang, Qiang and Tian, Zheng and Wang, Jingbo and Pang, Jiangmiao and Zhang, Weinan},
  journal={arXiv preprint arXiv:2507.07356},
  year={2025}
}

@article{liao2025beyondmimic,
  title={Beyondmimic: From motion tracking to versatile humanoid control via guided diffusion},
  author={Liao, Qiayuan and Truong, Takara E and Huang, Xiaoyu and Tevet, Guy and Sreenath, Koushil and Liu, C Karen},
  journal={arXiv preprint arXiv:2508.08241},
  year={2025}
}

@article{geyer2003positive,
  title={Positive force feedback in bouncing gaits?},
  author={Geyer, Hartmut and Seyfarth, Andre and Blickhan, Reinhard},
  journal={Proceedings of the Royal Society of London. Series B: Biological Sciences},
  volume={270},
  number={1529},
  pages={2173--2183},
  year={2003},
  publisher={The Royal Society}
}

@article{sreenath2011compliant,
  title={A compliant hybrid zero dynamics controller for stable, efficient and fast bipedal walking on MABEL},
  author={Sreenath, Koushil and Park, Hae-Won and Poulakakis, Ioannis and Grizzle, Jessy W},
  journal={The International Journal of Robotics Research},
  volume={30},
  number={9},
  pages={1170--1193},
  year={2011},
  publisher={SAGE Publications Sage UK: London, England}
}

@article{wang2025beamdojo,
  title={Beamdojo: Learning agile humanoid locomotion on sparse footholds},
  author={Wang, Huayi and Wang, Zirui and Ren, Junli and Ben, Qingwei and Huang, Tao and Zhang, Weinan and Pang, Jiangmiao},
  journal={arXiv preprint arXiv:2502.10363},
  year={2025}
}

@article{li2023robust,
  title={Robust and versatile bipedal jumping control through reinforcement learning},
  author={Li, Zhongyu and Peng, Xue Bin and Abbeel, Pieter and Levine, Sergey and Berseth, Glen and Sreenath, Koushil},
  journal={arXiv preprint arXiv:2302.09450},
  year={2023}
}

@article{huang2025learning,
  title={Learning humanoid standing-up control across diverse postures},
  author={Huang, Tao and Ren, Junli and Wang, Huayi and Wang, Zirui and Ben, Qingwei and Wen, Muning and Chen, Xiao and Li, Jianan and Pang, Jiangmiao},
  journal={arXiv preprint arXiv:2502.08378},
  year={2025}
}

@article{he2025learning,
  title={Learning getting-up policies for real-world humanoid robots},
  author={He, Xialin and Dong, Runpei and Chen, Zixuan and Gupta, Saurabh},
  journal={arXiv preprint arXiv:2502.12152},
  year={2025}
}

@inproceedings{li2025hold,
  title={Hold My Beer: Learning Gentle Humanoid Locomotion and End-Effector Stabilization Control},
  author={Li, Yitang and Zhang, Yuanhang and Xiao, Wenli and Pan, Chaoyi and Weng, Haoyang and He, Guanqi and He, Tairan and Shi, Guanya},
  booktitle={RSS 2025 Workshop on Whole-body Control and Bimanual Manipulation: Applications in Humanoids and Beyond}
}

@article{zhang2025falcon,
  title={FALCON: Learning Force-Adaptive Humanoid Loco-Manipulation},
  author={Zhang, Yuanhang and Yuan, Yifu and Gurunath, Prajwal and He, Tairan and Omidshafiei, Shayegan and Agha-mohammadi, Ali-akbar and Vazquez-Chanlatte, Marcell and Pedersen, Liam and Shi, Guanya},
  journal={arXiv preprint arXiv:2505.06776},
  year={2025}
}

@article{su2025hitter,
  title={Hitter: A humanoid table tennis robot via hierarchical planning and learning},
  author={Su, Zhi and Zhang, Bike and Rahmanian, Nima and Gao, Yuman and Liao, Qiayuan and Regan, Caitlin and Sreenath, Koushil and Sastry, S Shankar},
  journal={arXiv preprint arXiv:2508.21043},
  year={2025}
}

@article{peng2021amp,
  title={Amp: Adversarial motion priors for stylized physics-based character control},
  author={Peng, Xue Bin and Ma, Ze and Abbeel, Pieter and Levine, Sergey and Kanazawa, Angjoo},
  journal={ACM Transactions on Graphics (ToG)},
  volume={40},
  number={4},
  pages={1--20},
  year={2021},
  publisher={ACM New York, NY, USA}
}

@article{ho2022classifier,
  title={Classifier-free diffusion guidance},
  author={Ho, Jonathan and Salimans, Tim},
  journal={arXiv preprint arXiv:2207.12598},
  year={2022}
}

@article{zhou2025roborefer,
    title={RoboRefer: Towards Spatial Referring with Reasoning in Vision-Language Models for Robotics},
    author={Zhou, Enshen and An, Jingkun and Chi, Cheng and Han, Yi and Rong, Shanyu and Zhang, Chi and Wang, Pengwei and Wang, Zhongyuan and Huang, Tiejun and Sheng, Lu and others},
    journal={arXiv preprint arXiv:2506.04308},
    year={2025}
}

@article{zhou2025robotracer,
    title={RoboTracer: Mastering Spatial Trace with Reasoning in Vision-Language Models for Robotics},
    author={Zhou, Enshen and Chi, Cheng and Li, Yibo and An, Jingkun and Zhang, Jiayuan and Rong, Shanyu and Han, Yi and Ji, Yuheng and Liu, Mengzhen and Wang, Pengwei and others},
    journal={arXiv preprint arXiv:2512.13660},
    year={2025}
}
}


\end{document}